\documentclass[letterpaper, 10 pt, conference]{ieeeconf}

\IEEEoverridecommandlockouts

\usepackage{cite}
\usepackage{listings}
\usepackage{float}
\usepackage[dvipdfmx]{graphicx}
\usepackage{amssymb}
\usepackage{latexsym}
\usepackage{amsfonts}
\usepackage{url}
\usepackage{comment}

\newcommand{\FIG}[3]{
\begin{minipage}[b]{#1cm}
\begin{center}
\includegraphics[width=#1cm]{#2}\\
{\scriptsize #3}
\end{center}
\end{minipage}
}

\newcommand{\editage}[2]{#1}

\begin{document}

\newcommand{\figE}{
\begin{figure}[t]
\FIG{8.5}{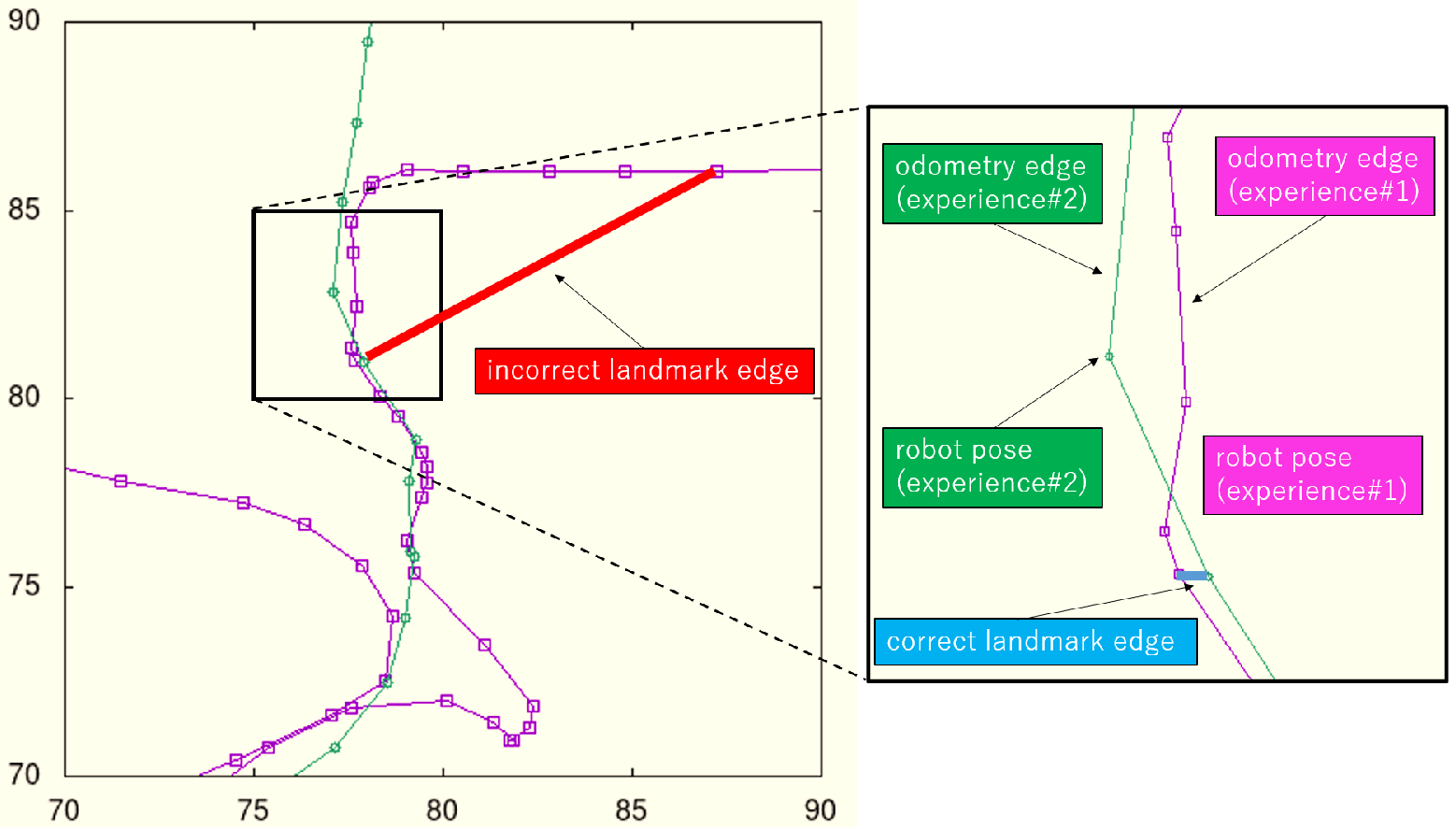}{}
\caption{Single-robot multi-session pose-graph SLAM.}\label{fig:E}
\end{figure}
}

\newcommand{\figC}{
\begin{figure}[t]
\vspace*{5mm}
\hspace*{-5mm}\FIG{8}{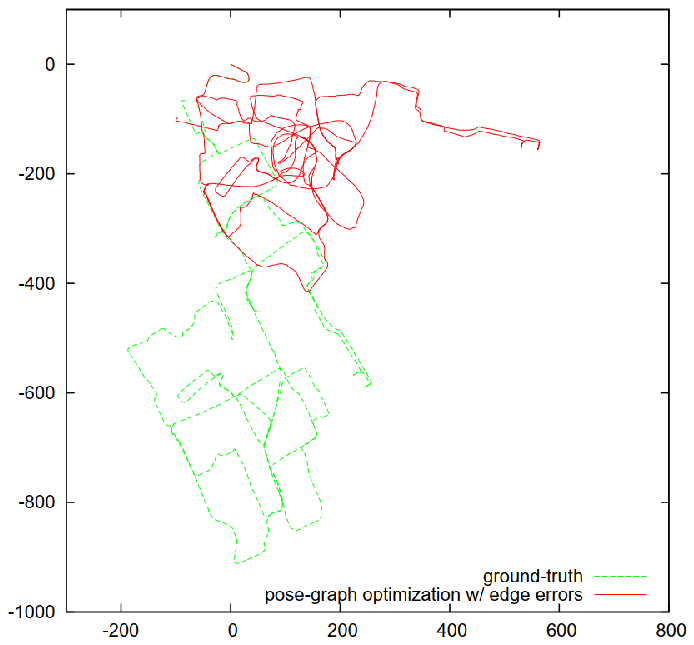}{}\vspace*{-5mm}\\
\caption{In this study, we present a unified framework for determining maximal consistent subsets in a given pose-graph map. As shown, errors in landmark edges can have catastrophic effects on the inferred pose-graph map. Detecting such landmark misrecognitions to avoid the catastrophic errors is the main focus of our study.}\label{fig:C}
\end{figure}
}

\newcommand{\figA}{
\begin{figure}[t]
\FIG{8.5}{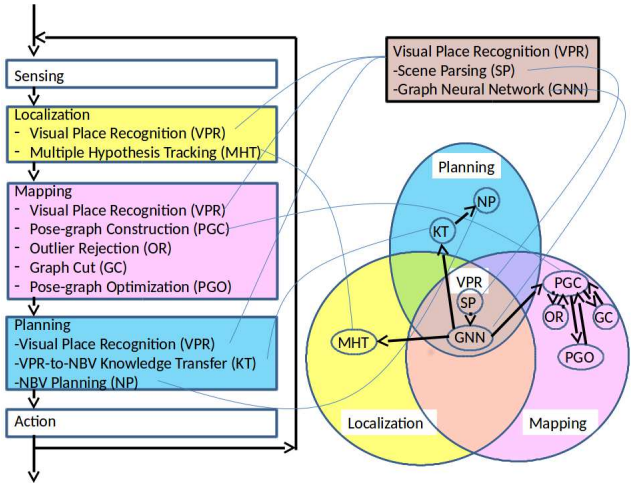}{}
\caption{Graph neural SLAM framework.}\label{fig:A}
\end{figure}
}

\newcommand{\figBs}{\hspace*{-10mm}}

\newcommand{\figB}{
\begin{figure*}[t]
\begin{center}
\FIG{17}{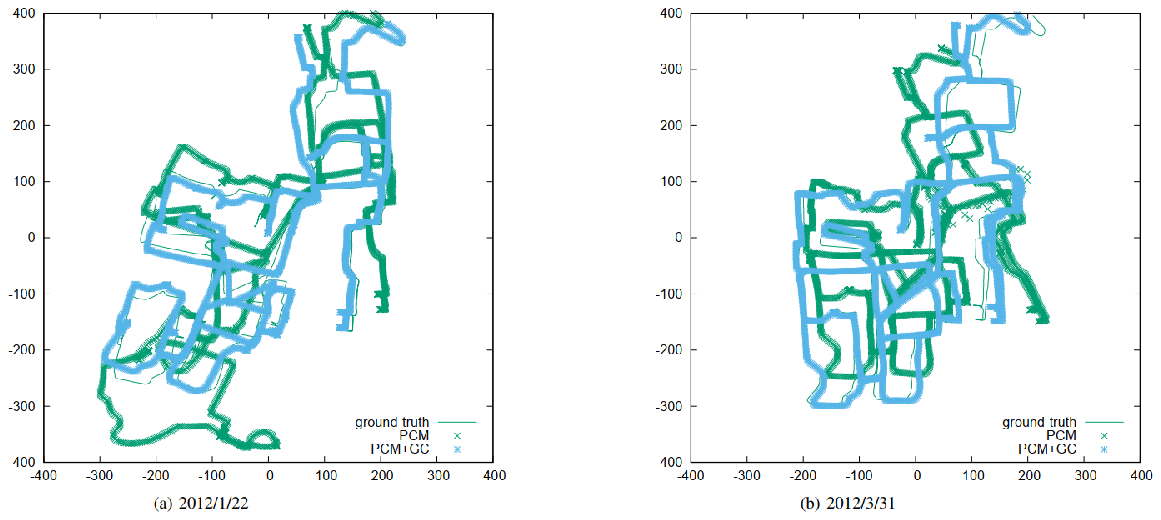}{}\\
\FIG{17}{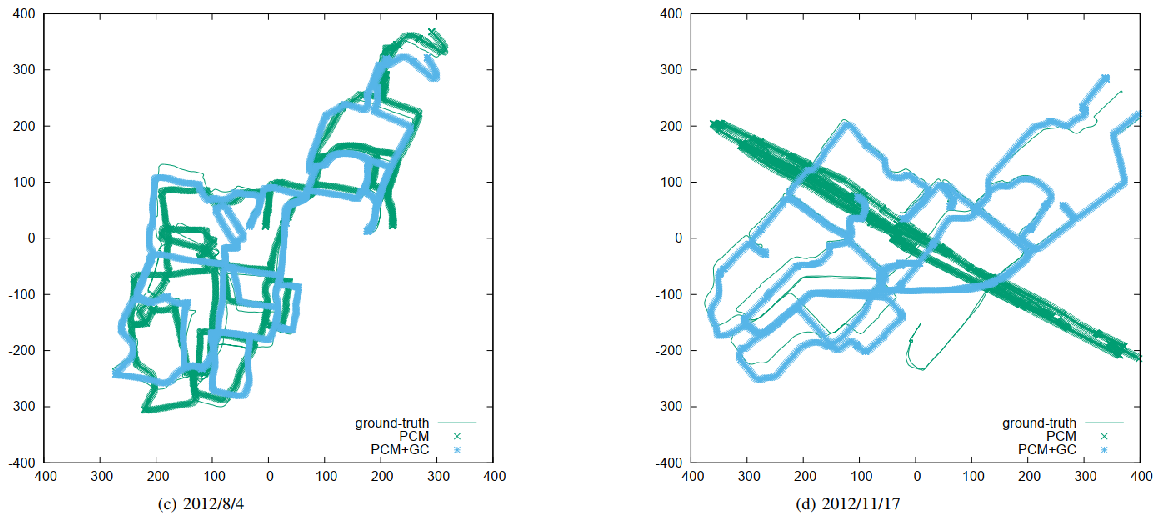}{}
\caption{Examples of pose-graph optimization.}\label{fig:5}
\end{center}
\end{figure*}
}

\newcommand{\tabA}{
\begin{table}[t]
\caption{Performance results.}\label{table:2}
\begin{tabular}{|r|r|r|r|r|}
\hline
& 2012/1/22 & 2012/3/31 & 2012/8/4 & 2012/11/17 \\
\hline
PCM & 97.4 & 72.7 & 33.5 & 214.6 \\
\hline
PCM+GC & 19.4 & 11.7 & 24.2 & 40.5 \\
\hline
\end{tabular}
\end{table}
}

\newcommand{\figD}{
\begin{figure}[t]
\centering
\FIG{8}{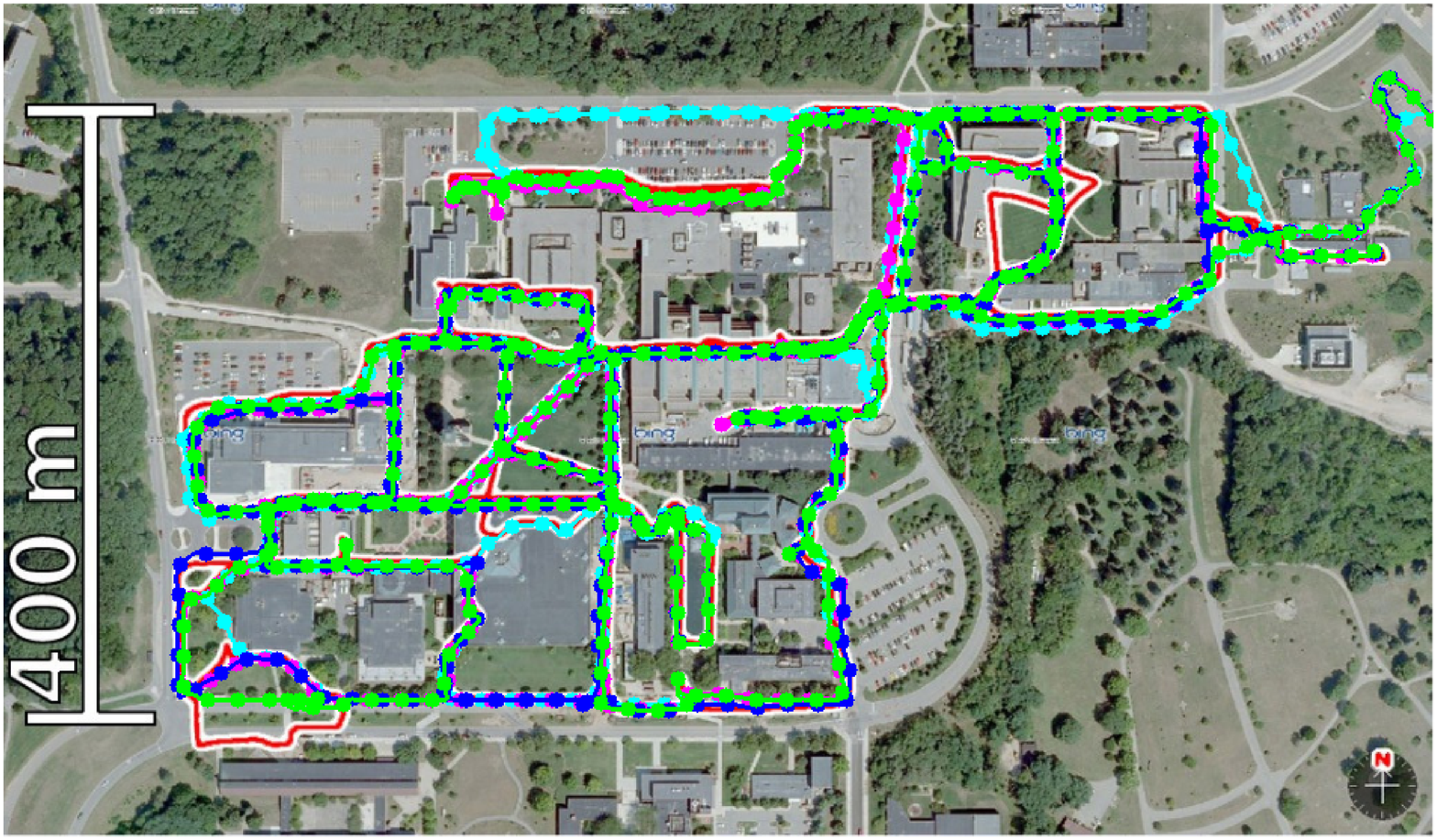}{}
\caption{Experimental environment. The trajectories of the four datasets, ``2012/01/22", ``2012/03/31", ``2012/08/04", and ``2012/11/17", used in our experiments are visualized in green, purple, blue, and light-blue curves and overlaid on the bird's eye view imagery obtained from the NCLT dataset.}\label{fig:D}
\end{figure}
}

\title{\LARGE \bf
Minimum Cost Multicuts
for Incorrect Landmark Edge Detection
in Pose-graph SLAM 
}

\author{%
\editage{Kazushi Aiba, Kanji Tanaka, and Ryogo Yamamoto
\thanks{Our work has been supported in part by JSPS KAKENHI Grant-in-Aid for Scientific Research (C) 20K12008 and 17K00361.}
\thanks{K. Aiba, K. Tanaka, and R. Yamamoto and are with Department of Engineering, University of Fukui, Japan. {\tt\small tnkknj@u-fukui.ac.jp}}%
}{TBA}}

\maketitle

\begin{abstract}
Pose-graph SLAM is the de facto standard framework for constructing large-scale maps from multisession experiences of relative observations and motions during visual robot navigations. It has received increasing attention in the context of recent advanced SLAM frameworks such as graph neural SLAM. One remaining challenge is landmark misrecognition errors (i.e., incorrect landmark edges) that can have catastrophic effects on the inferred pose-graph map. In this study, we present comprehensive criteria to maximize global consistency in the pose graph using a new robust graph cut technique. Our key idea is to formulate the problem as a minimum-cost multicut that enables us to optimize not only landmark correspondences but also the number of landmarks while allowing for a varying number of landmarks. This makes our proposed approach invariant against the type of landmark measurement, graph topology, and metric information, such as the speed of the robot motion. The proposed graph cut technique was integrated into a practical SLAM framework and verified experimentally using the public NCLT dataset.
\end{abstract}

\section{Introduction}

Pose-graph SLAM is the de facto standard framework for constructing large-scale maps from a sequence of relative landmark and odometry observations during visual robot navigations \cite{iSAM,DS4,PCM}. A pose-graph is a graph with the viewpoints of the robot as nodes and with two types of edges: landmark and odometry. An odometry edge represents the relative movement from the viewpoint of the robot at time $t$ to the next viewpoint at time $t+1$. A landmark edge represents the relative pose of a pair of viewpoints from which the identical landmark is observed. Given such a pose graph, pose-graph optimization is performed to determine an optimal pose configuration that explains all landmark and odometry edges with the smallest explanation error. The task is typically formulated as a nonlinear incremental optimization to address the incremental addition/deletion of pose nodes and landmark/odometry edges during the robot navigations \cite{iSAM}.

One remaining challenge is the misrecognition of landmark edges, which can have catastrophic effects on the inferred pose-graph map (Fig. \ref{fig:C}) \cite{DS4}. In a typical implementation that uses image features detected by an onboard camera as landmarks, landmark misrecognition often occurs \cite{FABMAP}. This landmark misrecognition causes incorrect landmark edges in the pose-graph map. Pose-graph optimization of such an incorrect pose graph results in catastrophic errors in the inferred pose-graph map. To address this problem, robust landmark recognition methods have been developed by several researchers. In \cite{FABMAP}, high-speed retrieval of landmark correspondence candidates using a bag-of-words model and subsequent geometric verification was presented. In 
\cite{island}, error removal was further performed via multiframe landmark correspondences. In 
\cite{PCM}, a criterion of pairwise consistency was introduced to detect errors from multiple robot experiences in a computationally tractable manner.

\editage{
\figC
}{}

In this study, we present a unified graph-cut (GC) framework for determining maximally consistent subsets in a given pose graph (Fig. \ref{fig:E}). Our key idea is to formulate the problem as a minimum-cost multicut \cite{MCMC}. Unlike the commonly used spectral clustering formulation, the minimum cost multicut formulation gives a natural rise to optimization not only for landmark correspondences but also for the number of landmarks while allowing for a varying number of landmarks. This is a desirable property for pose-graph applications, in which the number of valid landmarks is a-priori unknown. Specifically, we implemented a new consistency metric that relies only on the order of the observed landmarks, and that is invariant against the type of landmark measurement, graph topology, and metric information, such as the speed of the robot motion. Our approach for pose-graph robustification has the same format for input and output pose-graphs, and works with minimum assumptions on the error rate of landmark edges. Thus, it is orthogonal to the aforementioned outlier removal methods (e.g., \cite{FABMAP, island, PCM}) and can be combined with them to improve the overall performance. Experiments using the public NCLT dataset \cite{NCLT} validated that the proposed approach could boost the state-of-the-art method.

\figE

\editage{
\figA
}{}

\section{Graph Neural SLAM}

The graph neural SLAM framework is briefly introduced in this section (Fig. \ref{fig:A}). It is a deep learning extension of the scalable pose graph SLAM using the deep graph neural networks. The system comprises three modules: localization, mapping, and planning. The localization module localizes and tracks the robot's viewpoint using a multiple-hypothesis tracker. The mapping module builds and updates an environment map using pose graph construction and optimization. The planning module predicts the robot's next-best-view action from the recognized state of the robot's surroundings using a pre-trained state-to-action regressor. Notably, the visual place recognition (VPR) submodule is used commonly in all the aforementioned three modules for hypothesizing the viewpoint in localization, constructing landmark edges in mapping, and recognizing the state in planning. That is, the core of the SLAM system is the VPR module. In \cite{icra21takeda}, the development of robust VPR algorithm was explored by using the scene graph model and graph convolutional neural network, which could actually boost the state-of-the-art VPR. In the current study, we focus on the VPR task in the context of the mapping module and present a new method for mitigating the effects of landmark misrecognition in that module.

\section{Mapping Algorithm}

The proposed mapping algorithm (``Mapping" in Fig. \ref{fig:A}) consists of four steps. The pose-graph construction (``PGC") step acquires sensor data, updates landmark and odometry edges with landmark observations, searches for landmark correspondence, and extends the pose graph (Fig. \ref{fig:E}). The outlier rejection (``OR") step checks the local graph consistency to reject spurious landmark edges, which may be caused by perceptual aliasing. The graph cut (``GC") step, which we aim to introduce in this study, checks the global graph consistency to minimize the global inconsistency in the pose graph. The pose-graph optimization (``PGO") step searches for the optimal pose configuration that explains all landmark and odometry edges with the smallest explanation error. These steps are detailed in the following subsections, \ref{sec:construct}, \ref{sec:outlier}, \ref{sec:cut}, and \ref{sec:optimize}.

\subsection{Pose-graph Construction}\label{sec:construct}

The pose-graph construction step is responsible for estimating the robot's trajectory given a sequence of landmark and odometry edges. The trajectory of a robot is represented as a discrete set of poses, describing the position and orientation of the on-board camera at each keyframe. We denote the trajectory of the robot's $j-$th experience as $x$ = $[x_0^j, \cdots, x_i^j, \cdots ]$, where $x_i^j$ = $[R_i^j, t_i^j]\in$ SE(3). $R_i^j\in$ SO(3) and $t_i^j\in$ $\mathbb{R}^3$ respectively represent the rotation and translation of the pose associated with the  $i$-th viewpoint of the $j$-th experience.

\subsection{Outlier Rejection}\label{sec:outlier}

The outlier rejection step rejects spurious landmark edges, which may be caused by near-duplicate landmarks and by perceptual aliasing. We adopted the pairwise consistent measurement set maximization (PCM) technique proposed in \cite{PCM} for outlier rejection. The key insight behind PCM is to check whether or not pairs of landmark edge constraints are consistent with each other and then search for a large set of mutually consistent landmark edges (as shown in \cite{PCM}, the largest set of pairwise consistent measurements can be determined as a maximum clique). Although PCM does not check for the joint consistency of all measurements, the approach typically ensures that gross outliers are rejected. After pairwise consistency checks are performed, the maximum clique of the measurements for each of its neighbors is computed to determine inlier landmark edges.

\subsection{Pose-graph Cut}\label{sec:cut}

We formulated the problem of a pose-graph cut as an instance of a minimum-cost multicut \cite{MCMC}. Minimum cost multicut aims to decompose graph $G=(V, E)$ into the optimal number of segments to minimize the overall cost of edge weight $c_e$. This node-labeling problem can be formulated as a binary edge-labeling problem as follows:
\begin{equation}
\min_{y\in\{0, 1\}^E} \sum_{e\in E} c_e y_e 
\end{equation}
\[
\mbox{subject to $y\in$MC},
\]
where MC denotes a set of all the multicut characteristic functions of $G$. That is, all $y\in \{0,1\}^E$ form a closed boundary and represent a valid decomposition of the graph. Formally, these characteristic functions are described by cycle inequalities \cite{MCMC15}. To avoid trivial solutions, edge weights are usually chosen negative for edges that should be cut and positive for connected nodes that should be joined. 

We introduced the concept of {\it support edges} to evaluate the reliability of the edges of interest. The idea is that some grouping often exists in edge data for spatially similar pose nodes. For example, visually similar landmarks are often detected in images taken from spatially near viewpoints, which causes edges with spatially similar pose nodes. Importantly, this phenomenon often occurs only on the correct landmark edges and not on incorrect edges. Therefore, the presence or absence of such a group of mutually support edges can be used as an indicator of the correctness of each membership edge in the group.

The idea of using measurement groups for reliability evaluation is common in the field of robust estimation, such as RANSAC \cite{RANSAC}. In our new application, pose-graph cut, a support edge for an edge of interest consists of a pair of nodes with frame IDs that differ by at most $\Delta (=1)$ from the frame ID of one of the endpoint nodes of the edge of interest. Evidently, there were at most $4\Delta(\Delta+1)$ support edges. For a particular edge of interest, weight $c_e$ of that edge of interest is set to +1 if there is a support edge and to -1 otherwise.

\subsection{Pose-graph Optimization}\label{sec:optimize}

Pose-graph optimization was performed as a post-processing step of the aforementioned outlier removal and graph cuts. iSAM \cite{iSAM} was applied as a solver for pose-graph optimization because it is a general, easy-to-use, and incremental optimizer that allows the addition/removal of nodes/edges, making it suitable for map maintenance during real-time robot navigations.

\figD

\section{Experiments}

\figB

\tabA

We used the public NCLT dataset \cite{NCLT}. The NCLT dataset is a dataset collected by the Segway robot approximately every two weeks from January 8, 2012 to April 5, 2013 at the North Campus of the University of Michigan. Specifically, we used four image sequences from the NCLT dataset:  ``2012/1/22," ``2012/3/31," ``2012/8/4," and ``2012/11/17".
 Image data from front-facing camera and IMU data measured by the route shown in Fig. \ref{fig:D} were used in this study.

In the multi-experience scenario, we considered a simple setting in which the number of the experiences was two. Each of the four sequences was divided into two part of experience trajectories, in the first and the second halves. Then, two corresponding pose-graphs are computed from these two trajectories. Then, they are connected each other with correct landmark edges, which are defined by a pair of nodes that have spatially similar ground-truth poses. Then, a collection of incorrect edges is randomly added. The rate of incorrect edges over all landmark edges was empirically set to 10\%, considering the typical probability of landmark misrecognition caused in the graph neural SLAM framework \cite{icra21takeda}.

We found that there exist non-empty set of correct edges for all of the four datasets. Notably, such an overlapping part is often called ``revisit place" or ``loop closing" and plays a key role in robotic mapping \cite{LoopClosing}. Then, landmark correspondences were searched between each node of the submap, and all candidate correspondences were added as landmark edges to the pose-graph. To make the computational cost tractable, image frames in the original sequence in the NCLT dataset were down-sampled to 1/10.

\editage{

}{}

Figure \ref{fig:5} shows example results for pose-graph optimization with and without the proposed graph-cuts. Notably, the quality of the pose-graph could be actually improved by using the proposed method, and the resulted pose-graph was often similar to the ground-truth pose-graph.

For performance evaluation, the root mean square error (RMSE) over all poses in a pose graph of interest was used. First, a pseudo ground-truth pose-graph was computed for each pose-graph in the dataset by sending the input sequence without incorrect edges to a pose-graph optimization process, for which the same PCM software as in Section \ref{sec:optimize} was reused. Then, an inferred pose graph was evaluated in terms of the minimum RMSE over all possible coordinate transformations between the input and the ground-truth pose graphs.

Table \ref{table:2} lists the RMSE errors [m] with the correct map data for each of the results with and without the proposed graph-cuts. 
The results confirmed that the RMSE of the PCM method was 
104.6 m on average 
and that of the proposed method was 
	23.9 m, 
which was more than
four times higher in accuracy.

\section{Conclusions}

In this study, the problem of detecting landmark misrecognition in multisession pose-graph SLAM was reformulated as a minimum-cost multicut task. Our approach for pose-graph robustification had the same format for input and output pose-graphs and worked with minimum assumptions on the error rate of landmark edges. Thus, it is orthogonal to the existing methods for outlier removal (e.g., PCM \cite{PCM}) and can be combined with them to improve the overall performance. Experiments using the public NCLT dataset validated that the proposed approach could boost the state-of-the-art method.

\bibliography{reference} 
\bibliographystyle{unsrt} 

\end{document}